\documentclass[runningheads]{llncs}

\usepackage{eccv}

\usepackage{eccvabbrv}

\usepackage{graphicx}
\usepackage{booktabs}

\usepackage[accsupp]{axessibility}  

\usepackage{multirow, multicol}
\usepackage{enumitem}

\newcounter{alphaSection}
\renewcommand{\thealphaSection}{\Alph{alphaSection}}

\newcommand{\alphaSection}[1]{%
  \refstepcounter{alphaSection}%
  \section*{\thealphaSection\quad#1}%
  \addcontentsline{toc}{section}{\thealphaSection\quad#1}%
}

\usepackage{hyperref}

\usepackage{orcidlink}

\begin{document}

\title{FALCON: Frequency Adjoint Link with CONtinuous Density Mask for Fast Single Image Dehazing} 

\titlerunning{FALCON}

\author{
Donghyun Kim
\quad
Seil Kang
\quad
Seong Jae Hwang\thanks{Corresponding author.}
}

\authorrunning{Kim et al.}

\institute{Yonsei University\\ 
\email{\{danny0103, seil, seongjae\}@yonsei.ac.kr}}

\maketitle

\begin{abstract}
    Image dehazing, addressing atmospheric interference like fog and haze, remains a pervasive challenge crucial for robust vision applications such as surveillance and remote sensing under adverse visibility. While various methodologies have evolved from early works predicting transmission matrix and atmospheric light features to deep learning and dehazing networks, they innately prioritize dehazing quality metrics, neglecting the need for real-time applicability in time-sensitive domains like autonomous driving. This work introduces FALCON (Frequency Adjoint Link with CONtinuous density mask), a single-image dehazing system achieving state-of-the-art performance on both quality and speed. Particularly, we develop a novel bottleneck module, namely, Frequency Adjoint Link, operating in the frequency space to globally expand the receptive field with minimal growth in network size. Further, we leverage the underlying haze distribution based on the atmospheric scattering model via a Continuous Density Mask (CDM) which serves as a continuous-valued mask input prior and a differentiable auxiliary loss. Comprehensive experiments involving multiple state-of-the-art methods and ablation analysis demonstrate FALCON's exceptional performance in both dehazing quality and speed (\ie, $>$180 frames-per-second), quantified by metrics such as FPS, PSNR, and SSIM.
\end{abstract}    
\section{Introduction}
\label{section:introduction}

Eliminating naturally occurring visual distortions from atmospheric interference such as fog, smoke, and haze from an image, commonly termed \textit{image dehazing}, is a ubiquitous task in vision applications. For instance, image dehazing plays a pivotal role in enhancing the visual quality of images captured in adverse weather conditions, providing clearer insights into fields such as surveillance~\cite{salazar2018fast, ullah2021light}, autonomous driving~\cite{mehra2020reviewnet, wang2022cycle}, and remote sensing. Despite its practical significance, image dehazing faces inherent challenges, including the variability of atmospheric conditions, the complex interaction of light with particles, and the delicate balance between preserving image details and reducing haze artifacts.

\begin{figure}[t]
    \includegraphics[width=\linewidth]{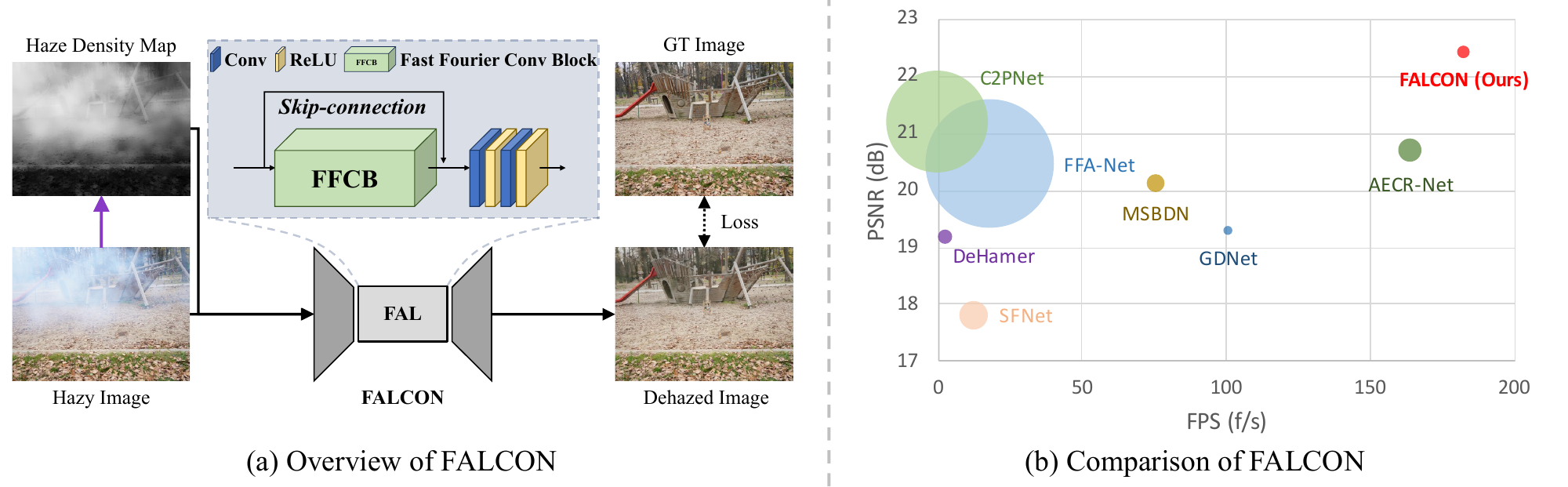}
    \caption{(a) A Simplified Illustration of the FALCON Workflow. The purple arrow represents the operation of calculating the haze density map. We implement this process through an approach called Differentiable Density Pooling. (b) Analysis of dehazing performance (PSNR, Peak Signal-to-Noise Ratio) vs. dehazing speed (FPS, frames-per-second) on NH-Haze2 dataset with images of size 256$\times$256 using RTX 3090 GPU. For each method, its circle size is proportional to the FLOPS. The goal is to achieve both high PSNR for quality and FPS for speed. Our method FALCON achieves the highest PSNR (22.41 dB) with the fastest inference FPS (182.90 frames per second), enabling a real-time single image dehazing while achieving the best dehazing quality.}
    \label{fig:plot}  
\end{figure}

In response to the aforementioned challenges in image dehazing, numerous methodologies have emerged to tackle the challenges presented by image degradation. Early research~\cite{dcp, fattal2008single, tan2008visibility, zhu2015fast, fattal2014dehazing} predominantly concentrated on predicting features such as the transmission matrix and atmospheric light, relying on various haze-related priors and assumptions. Concurrently, the rise of deep learning has spurred the development of various dehazing networks~\cite{dehazenet, gfn, gdnet, msbdn, aecr}. With the progression of diverse research progress, the field of single image dehazing has expanded, witnessing a continuous influx of high-performing methods for haze removal.

While existing dehazing methods demonstrate great performance, they primarily focus on metrics like PSNR and SSIM, which measure the quality of dehazed images. However, in various practical applications such as autonomous driving, CCTV, and national defense technology where images must be dehazed quickly for responsive subsequent actions, the dehazing \textit{speed} is another significant and perhaps necessary condition to pursue. 
Yet, existing state-of-the-art dehazing methods, primarily network-based, fall short in achieving \textit{real-time applicability} where each image is dehazed within milliseconds.

This work aims to deliver a single image dehazing system that meets the demands of both quality and speed (\cref{fig:plot}(b)).
Specifically, we focus on developing a network-based dehazing model, which the current top-performing methods consistently rely on. In that sense, to achieve fast inference time, our methodological efforts make minimal architectural changes to preserve the computational cost and concentrate on building a simple learning framework (\cref{fig:plot}(a)). We show how even a simple vanilla U-Net~\cite{unet} outperforms existing deep image dehazing models with the following methodological contributions.

The simple structure of a U-Net~\cite{unet} lacks depth, posing challenges in achieving a sufficiently wide receptive field through conventional convolution operations for handling global image features. Also, real-world images often have high resolutions, such as 1600$\times$1200, necessitating deeper networks at the expense of increased computational costs. To obtain larger receptive fields without proportionally growing the model, we introduced the Frequency Adjoint Link (FAL), inspired by the Fast Fourier Convolution~\cite{ffc}. This lightweight replacement of dense CNN-based bottlenecks facilitates comprehensive information capture from the image by performing convolution operations not only in the spatial domain but also in the frequency domain. 

Although recent methods primarily leverage powerful deep models, various preceding works utilized a more fundamental understanding of haze as a type of atmospheric interference, namely, the atmospheric scattering model. The most prominent work, Dark Channel Prior~\cite{dcp}, showed how a simple assumption about haze led to characterizing haze with a transmission map. Our work uniquely views the density map as a prior functioning as a continuous-valued mask, named Continuous Density Mask (CDM), indicating the degree of image degradation. Remarkably, a simple concatenation of this density-based prior yields potent cues about the haze distribution, significantly enhancing the learning process. Notably, we leverage CDM not only as the prior but also as an auxiliary loss by deriving a differentiable form of the density estimation.

\noindent\textbf{Contributions.} In this work, we introduce \textit{FALCON} (\textbf{F}requency \textbf{A}djoint \textbf{L}ink with \textbf{CON}tinuous density mask) for fast single image dehazing.
Specifically, we make the following contributions:

\begin{itemize}
	\item We propose a novel bottleneck module called Frequency Adjoint Link (FAL) designed to efficiently operate in the frequency space, effectively expanding the receptive field with minimal architectural growth.
	
	\item We also cleverly utilize Continuous Density Mask (CDM) characterizing the haze distribution based on the atmospheric scattering model. CDM is extremely efficiently computed, ensuring minimal computational burden, and is seamlessly integrated as a potent input prior and a differentiable auxiliary loss.
 
	\item Our study presents comprehensive results from diverse experiments, including comparisons with other state-of-the-art methods and an ablation analysis. Through metrics such as FPS, PSNR, and SSIM, we illustrate that our method achieves state-of-the-art performance, delivering exceptional speed alongside impressive dehazing quality.
\end{itemize}

We provide the code in the supplementary material which will be released upon publication.
\section{Related Works}
\label{section:related_works}

\noindent\textbf{Atmospheric Scattering Model.} 
In the image dehazing literature, various methodologies~\cite{tan2008visibility, zhu2015fast, narasimhan2000chromatic, narasimhan2002vision} have interpreted haze-like image degradation using an atmospheric scattering model~\cite{mccartney1976optics, nayar1999vision, narasimhan2003contrast} formulated as follows:
\begin{equation} \small
\textbf{I}(x) = \textbf{J}(x)t(x) + \textbf{A}(1-t(x)),
\end{equation}
where $\textbf{I}$ is the hazy image, $\textbf{J}$ is the clear or ground truth image before distortion, $\textbf{A}$ is the global atmospheric light, and $x$ is the index of pixel. Of particular note, $t(x)\in(0,1)$ is the medium transmission characterizing the proportion of light that reaches the camera after passing through the atmosphere. Conversely speaking, the proportion of light that does \textit{not} reach the camera due to scattering is represented as $1 - t(x)$, which is also interpreted as the \textit{haze density}.

\noindent\textbf{Prior-informed Dehazing.} Such finding has led early works to explicitly focus on predicting features like the transmission map and atmospheric light using haze-related priors and assumptions~\cite{dcp, fattal2008single, tan2008visibility, zhu2015fast}. A notable example is the Dark Channel Prior (DCP)~\cite{dcp} which often was combined with the atmospheric scattering model to effectively facilitate the transmission map computation, significantly advancing the field of dehazing. Similar efforts have also appeared including maximizing local contrast for haze removal and predicting the scene's albedo~\cite{tan2008visibility, fattal2008single}.

\noindent\textbf{Network-based Dehazing.} Deep networks also began showing promising results. Starting from MSCNN~\cite{mscnn} and DehazeNet~\cite{dehazenet}, various deep models have attempted to predict the medium transmission map or the final haze-free image using CNNs, achieving significant improvements~\cite{dcpdn, li2018single, ppdnet, mei2019progressive, engin2018cycle}. In particular, recent methods like DeHamer~\cite{dehamer} have proposed combining CNNs and transformers to harness the strengths of both. Meanwhile, methods like C2PNet~\cite{c2p} have achieved impressive results using a novel learning approach called contrastive regularization.

\noindent\textbf{Frequency Domain in Dehazing.} In real-world haze images, the haze frequently appears as a \textit{global} artifact, covering the majority of the image region. However, prior networks with convolutions have innate drawbacks of having a limited receptive field, making it challenging to efficiently derive global features. In response, inspired by prior vision techniques~\cite{yu2015multi, yu2018generative, wang2018image, ffc}, several dehazing methods operate in the frequency domain in which the corresponding Fourier bases span the whole image. For example, DW-GAN~\cite{dwgan} proposed a two-branch network using wavelet transform, and many other studies~\cite{yang2019wavelet, liu2018multi, wang2021ensemble, mao2023intriguing} have proposed methodologies that apply the frequency domain using wavelet and Fourier transforms. Our work utilizes the Fast Fourier Convolution~\cite{ffc} to apply the convolution in the frequency domain, effectively expanding the receptive field in a simple convolutional framework.
\section{Methods}
\label{section:methods}

\begin{figure}[t]
    \center
    \includegraphics[width=\linewidth]{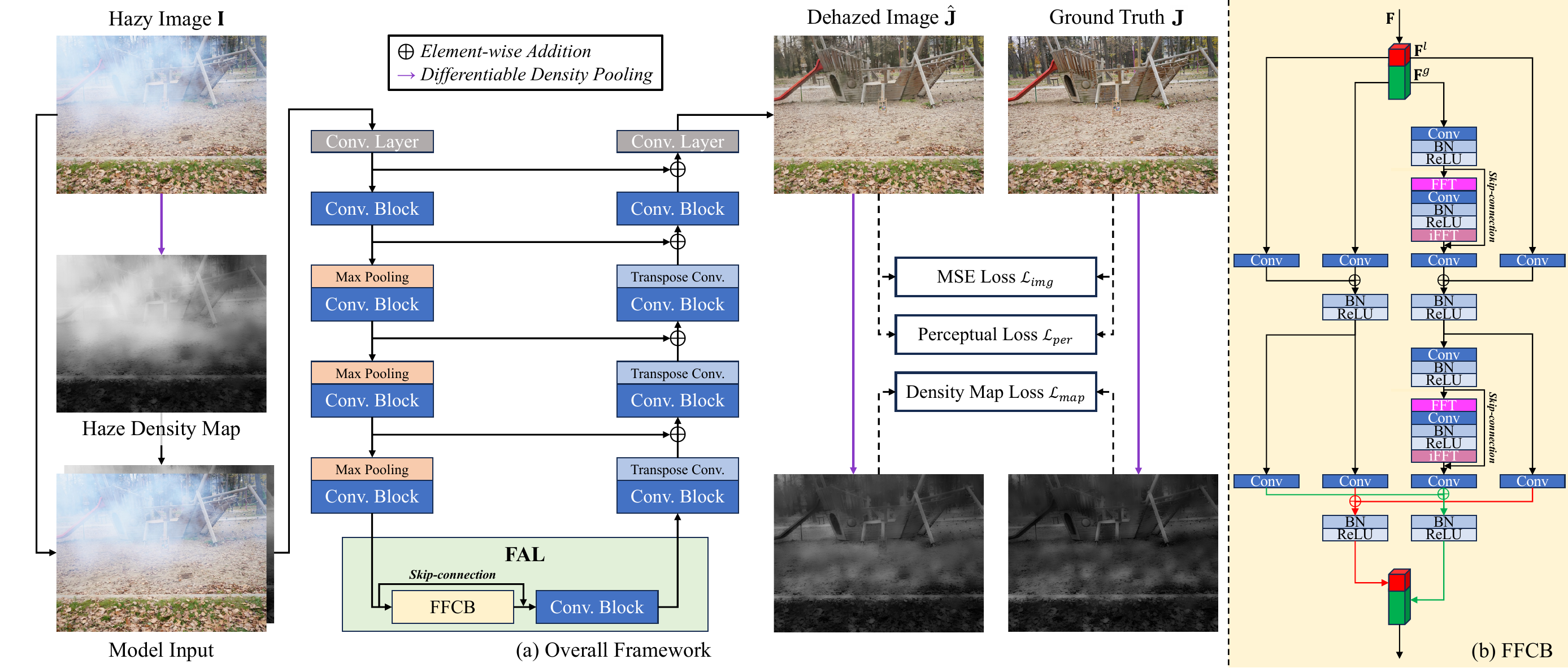}
    \caption{Overall pipeline of our single image dehazing method, FALCON. (1) The input hazy image is concatenated with its haze density map, namely, Continuous Density Mask, as a haze prior. (2) Our main network takes the concatenated input and leverages our Frequency Adjoint Link (FAL) to efficiently exploit the frequency domain. (3) The output dehazed image is compared against the ground-truth image in the image space ($\mathcal{L}_{img}$), density map space ($\mathcal{L}_{map}$), and VGG-16 feature space ($\mathcal{L}_{per}$).}
    \label{fig:pipeline}    
\end{figure}

In this section, we describe our method, FALCON, as shown in \cref{fig:pipeline}. First, we briefly outline the algorithmic process. Then, in detail, we cover each technical contribution of our approach.

\subsection{FALCON: Overview}
Given an input 3-channel hazy image $\textbf{I} \in \mathbb{R}^{H \times W \times 3}$, the goal is to generate a 3-channel dehazed image $\hat{\textbf{J}}\in \mathbb{R}^{H \times W \times 3}$ which is close to the ground-truth clear image $\textbf{J}\in \mathbb{R}^{H \times W \times 3}$. In line with this general framework, our method brings the following novel contributions.
(1) First, our model uniquely computes the haze density map, namely, Continuous Density Mask (CDM), based on the principle from the Dark Channel Prior (DCP)~\cite{dcp}. In particular, the original density map calculation involves a nested minimization which is seemingly non-differentiable and CPU-intensive. However, we identify that this process easily equates to a special type of pooling, turning the density map calculation to be an extremely computationally efficient process. (2) Next, we introduce our novel Frequency Adjoint Link (FAL) module which greatly improves the vanilla U-Net~\cite{unet} by efficiently expanding the receptive field with minimal network size increase. (3) Lastly, our differentiable density map estimation allows us to incorporate a new kind of loss function based on the density map. Next, we describe each of the components in detail.

\subsection{Frequency Adjoint Link}
For dense haze images which our study specifically considers, the existing networks often struggled with regions with extensive haze. When a significant area exhibits such nature, the corresponding areas showed poor dehazing results. We have identified the root of this issue as the limited receptive field of the convolution used in the network, not being able to consider the relatively clear regions beyond the large hazy areas. In our work, to achieve dehazing with a broader receptive field, we leverage the frequency domain. In the spatial domain, each pixel of the image only has a value corresponding to its position, dealing with very local information. However, in the frequency domain, each point represented on a pixel contains information about the signal that constitutes the entire image. Therefore, performing convolution in the frequency domain allows for a wide receptive field without using many pixels ~\cite{katznelson2004introduction}. For the transition to the frequency domain, we utilize the Fourier transform and adeptly apply the concept of Fast Fourier Convolution~\cite{ffc} to develop a module called the \textit{Frequency Adjoint Link} (FAL).

We now describe the architecture of FAL in detail. Recall that the goal of FAL is to capture the haze pattern which is densely spread within an image. Specifically, it aims to consider both the \textit{local} haze pattern which may not be sufficiently captured with typical CNN receptive fields, and the \textit{global} haze pattern beyond the locally hazy area of interest. 

The FAL consists primarily of the Fast Fourier Convolution Block (FFCB) and convolution block, as shown in \cref{fig:pipeline}(a). Let $\mathbf{F} \in \mathbb{R}^{H\times W\times C}$ be the downsampled feature at the bottom-most layer of U-Net, fed into the FFCB. Then, as shown in \cref{fig:pipeline}(b), $F$ is first channel-wise partitioned as follows: $\mathbf{F} \rightarrow \{\mathbf{F}^l, \mathbf{F}^g \}$ where $\mathbf{F}^l\in\mathbb{R}^{H\times W\times (1-\alpha_{in})C}$ is the first $(1-\alpha_{in})C$ channels dedicated to the local route (\eg, spatial domain), and $\mathbf{F}^g\in\mathbb{R}^{H\times W\times \alpha_{in}C}$ is the next $\alpha_{in}C$ channels dedicated to the global route (\eg, frequency domain). $\alpha_{in}$ is the ratio of channels between $\mathbf{F}^l$ and $\mathbf{F}^g$. Specifically, $\mathbf{F}^g$ is transformed into the Fourier feature in the frequency domain via convolution block and Fast Fourier Transform (FFT) layer: $\mathbf{F}_{freq}^g=FFT(Conv(\mathbf{F}^g))$, where $Conv(\cdot)$ is the convolution block and $FFT(\cdot)$ denotes FFT layer. Then, the subsequent convolution is applied onto $\mathbf{F}_{freq}^g$ to derive the globally convolved receptive field. The resulting feature is then transformed back to the spatial domain via inverse Fast Fourier Transform (iFFT). 
More detailed information about the Fast Fourier Convolution Block can be found in the supplementary material. 

By adding just this Frequency Adjoint Link to the middle of the U-Net~\cite{unet}, we implemented an image dehazing network, which shows both speed and impressive dehazing performance.

\subsection{Continuous Density Mask}
Unlike some homogeneous types such as fog which almost uniformly covers the entire image with moderate visibility, we consider much more difficult cases of non-homogeneous and dense haze. That is, we observe dense haze formation resulting in very low visibility in almost all areas (\ie, Dense-Haze dataset~\cite{dense-haze}; \cref{fig:Dense_Haze} first column). Also, non-homogeneous formation of haze may also result in barely visible areas (\ie, NH-Haze and NH-Haze2 datasets~\cite{nh-haze, nh-haze2}; \cref{fig:NH_Haze2} first column). From the methodological perspective, such dense and/or non-homogeneous haze scenarios bring a much greater challenge. While physics-informed methods such as the Dark Channel Prior (DCP)~\cite{dcp} can cleverly estimate the haze density, we point out that the actual dehazing process requires a separate line of effort. Nonetheless, such haze density maps become a strong prior for the dehazing model, and we explicitly leverage this as input prior called the \textit{Continuous Density Mask} (CDM). We next describe how CDM is derived and utilized within our framework.

\subsubsection{Mask Mechanism}
In tasks like inpainting, where a mask is utilized throughout the pipeline, the mask typically assumes binary values of 0 or 1. This indicates whether a pixel retains its original value or is entirely lost. Borrowing this mechanism, we incorporated the concept of a mask in the dehazing task. Upon examining hazy images, especially real-world hazy images, it's evident that haze manifests as a complex distribution with varying densities across different locations. We interpreted the hazy image as a degraded version of the ground truth image, where each pixel is compromised to varying extents. Unlike inpainting, where compromised pixels lose all their values and assume a value of 0, in hazy images, pixels are degraded by the haze, not entirely lost. This means degradation of each pixel cannot be simply categorized into two values, 0 or 1. Instead, we assign a continuous value to each pixel based on the haze's density.

\subsubsection{Haze Density Calculation}
\label{section:DDP}
We utilized the Dark Channel Prior (DCP)~\cite{dcp} to compute the haze density map. DCP is based on the observation that most local patches in outdoor haze-free images contain at least one pixel which has a very low intensity in at least one color channel. Based on the atmospheric scattering model~\cite{mccartney1976optics, nayar1999vision, narasimhan2003contrast}, given a clean image $J$, its dark channel value $J^{dark}(x)$ at the pixel location $x$ is
\begin{equation} \small
J^{dark}(x) = \min_{c \in \{r, g, b\}}(\min_{y \in \Omega(x)}(J^c(y))),
\end{equation}
where $c$ is one of the $\{r, g, b\}$ channels, $y$ is a pixel from the local patch $\Omega(x)$ centered at pixel $x$. Thus, $J^c(y)$ is the value of channel $c$ of a neighboring pixel $y$, and $J^{dark}(x)$ is the minimum intensity value across all three channels within the window $\Omega(x)$. Then, following DCP, we assume that the dark channel value $J^{dark}(x)$ is very low, that is, $J^{dark}(x)=0$. Assuming the value of the global atmospheric light \( A \) as 1, the atmospheric scattering model~\cite{mccartney1976optics, nayar1999vision, narasimhan2003contrast}'s equation can be simplified using Dark Channel Prior~\cite{dcp} as
\begin{equation} \small
1 - t(x) = \min_{c \in \{r, g, b\}}(\min_{y \in \Omega(x)}(I^c(y))).
    \label{eq:density}
\end{equation}
As previously mentioned, \( 1 - t(x) \) represents the haze density map. We have implemented this operation in a differentiable manner, referring to it as \textit{Differentiable Density Pooling} (DDP). To perform the dual minimum operation present in \cref{eq:density}, we adapted what is known as Min Pooling. Differentiable Density Pooling is as follows:
\begin{equation} \small
1 - t(x) = DDP(I(x)) = -\mathcal{M}((\mathcal{M}(-I(x)))^T).
    \label{eq:DDP}
\end{equation}
$\mathcal{M}(\cdot)$ and \(DDP(\cdot)\) represents Max Pooling and Differentiable Density Pooling, respectively. While standard Min Pooling identifies local minimum values, our implementation goes a step further to ensure that the results are consistent with the existing density map calculations of \cref{eq:density}. Although both operations yield the same outcome, naively implementing the dual minimum operation from \cref{eq:density} would result in a non-differentiable process. However, our approach, the Differentiable Density Pooling, allows for a differentiable implementation that does not break the computational graph within the overall backpropagation. 
Also, DDP allows us to quickly calculate the haze denisty map using a DCP-based algorithm without using a trainable network.
Using \cref{eq:DDP}, We swiftly and seamlessly connect the haze density map for each pixel in the hazy image as an auxiliary channel with the input hazy image. Additionally, by utilizing Differentiable Density Pooling, we have also implemented our novel loss term.

\subsection{Loss Formulation}
Our dehazing framework optimizes a combination of Mean Squared Error (MSE) loss, Perceptual loss, and the newly introduced Density Map Loss to ensure high-quality image reconstruction.

\noindent\textbf{Mean Squared Error (MSE) Loss.} The MSE loss computes the pixel-wise accuracy between the dehazed image \( \hat{J} \) and its ground truth \( J \): 
\begin{equation}
\small
    \mathcal{L}_{img} = \frac{1}{N} \sum_{i=1}^{N} (\hat{J}(i) - J(i))^2.
\end{equation}
MSE provides a stable optimization landscape, aiding in model convergence and ensuring high-fidelity reconstructions.

\noindent\textbf{Perceptual Loss \cite{perc}.} This loss is a combination of concept and style losses, derived using a pre-trained VGG-16 network. The concept loss captures high-level semantic differences, while the style loss ensures stylistic fidelity. Specifically, for a given dehazed image \( \hat{J} \) and its ground truth \( J \),
\begin{equation} 
\small
\mathcal{L}_{concept} = || \phi_8(\hat{J}) - \phi_8(J) ||^2,
\end{equation}
\begin{equation} 
\small
\mathcal{L}_{style} = \sum_{l \in \{3, 8, 15\}} || G(\phi_l(\hat{J})) - G(\phi_l(J)) ||^2,
\end{equation}
where \( \phi_l \) represents the feature maps from the \( l^{th} \) layer of the VGG-16 network, and \( G \) denotes the Gram matrix. The overall Perceptual loss is then 
\begin{equation}
\small
\mathcal{L}_{per} = \mathcal{L}_{concept} + \mathcal{L}_{style}.    
\end{equation}

\noindent\textbf{Density Map Loss.} To further refine the dehazing process, we introduce the Density Map Loss \( \mathcal{L}_{map} \), which leverages the Differentiable Density Pooling \( DDP(\cdot) \) described in the previous subsection. This loss function measures the MSE between the haze density maps of the dehazed image \( DDP(\hat{J}) \) and the ground truth \( DDP(J) \), providing a direct optimization target for haze density accuracy: 
\begin{equation}
\small
    \mathcal{L}_{map} = \frac{1}{N} \sum_{i=1}^{N} (DDP(\hat{J})(i) - DDP(J)(i))^2.
\end{equation}

\noindent\textbf{Combined Objective.} The final loss function that we optimize is a weighted combination of the MSE, Perceptual, and Density Map losses:
\begin{equation} 
\small
\mathcal{L}_{final} = \alpha \mathcal{L}_{img} + \beta \mathcal{L}_{per} + \gamma \mathcal{L}_{map}
\end{equation}
where $\alpha>0$, $\beta>0$, and $\gamma>0$ are the weights. In the later section, we empirically evaluate the significance of using all three losses with an ablation study as well.
\section{Experiments}
\label{section:experiments}

We first describe the experimental setup such as datasets, implementation details, and baselines. Then, we present the experimental results on multiple real-world image dehazing datasets against existing state-of-the-art methods along with the ablation studies. Lastly, we extensively analyze the computational speed of various dehazing networks and demonstrate the versatility of Continuous Density Mask.

\subsection{Experimental Setup}
\noindent\textbf{Datasets.} Our experimentation involves three real-world datasets, and, consistent with previous studies~\cite{c2p, dehamer, aecr}, we conducted the experiments according to the official data splits of each dataset. (1) \textit{Dense-Haze}~\cite{dense-haze} dataset is known for its dense, homogeneous haze conditions. (2) \textit{NH-Haze}~\cite{nh-haze} dataset shows dense and non-homogeneous haze patterns. (3) Lastly, \textit{NH-Haze2}~\cite{nh-haze2} dataset is comprised of strong non-homogeneous hazy scenes. Additionally, for experiments on synthetic dataset, we employed the RESIDE~\cite{reside}, and more detailed information about these experimental results and the overall dataset can be found in the supplementary material.

\noindent\textbf{Implementation Details.}
The experiments involve various GPUs to evaluate both the dehazing quality and the computational efficiency of our method. Specifically, we used an NVIDIA RTX A6000 GPU for the PSNR and SSIM evaluations. To measure the dehazing speed in frames-per-second (FPS) across different resolutions and GPU settings, we employed a diverse set of NVIDIA GPUs: RTX A6000, RTX 3090, GTX 1080 Ti, and T4. Each inference time was measured and averaged over the corresponding test set samples. We used the Adam optimizer with an initial learning rate of 0.0001 with the random flipping and random cropping augmentations. The batch size was set to 5 for training.

\noindent\textbf{Baseline Methods.}
We compare our model FALCON to various traditional and recent state-of-the-art dehazing techniques. These include traditional methods like DCP~\cite{dcp}, which relies on the dark channel prior, and a variety of deep learning approaches such as DehazeNet~\cite{dehazenet}, AOD-Net~\cite{aod}, GCANet~\cite{gca}, GDNet~\cite{gdnet}, FFA-Net~\cite{ffa}, and MSBDN~\cite{msbdn}. Additionally, we have included recent advancements like AECR-Net~\cite{aecr}, Dehamer~\cite{dehamer}, SFNet~\cite{sfnet}, and C2PNet~\cite{c2p} to cover a broad spectrum of methodologies. 

\noindent\textbf{Evaluation Metrics.}
Our metrics include the conventional Peak Signal-to-Noise Ratio (PSNR) and Structural Similarity Index Measure (SSIM) which evaluate the dehazing quality. Additionally, we measure the Frames Per Second (FPS), quantifying the number of images dehazed by the model per second, providing insights into the actual dehazing speed for real-time applications. Lastly, we also compute FLoating point Operations Per Second (FLOPs) which quantifies the number of floating point operations of the model for a single input during inference. Thus, an ideal method should achieve high PSNR, SSIM, and FPS with low FLOPs.

\begin{table}[t]
    \caption{Quantitative comparisons against other methods on real-world datasets. FPS and FLOPs are based on the inference using an RTX 3090 GPU with images of 256x256 resolution.}
    \centering
    \scalebox{0.77}{
    \begin{tabular}{c c c c c c c c c c}
        \toprule
        \multirow{2}{*}{\centering Method} & \multirow{2}{*}{\centering Venue} & \multicolumn{2}{c}{Dense-Haze} & \multicolumn{2}{c}{NH-Haze} & \multicolumn{2}{c}{NH-Haze2} & \multicolumn{2}{c}{Computational Efficiency} \\
        
        \cmidrule(lr){3-4} \cmidrule(lr){5-6} \cmidrule(lr){7-8} \cmidrule(lr){9-10}
        
        & & PSNR $\uparrow$ & SSIM $\uparrow$ & PSNR $\uparrow$ & SSIM $\uparrow$ & PSNR $\uparrow$ & SSIM $\uparrow$ & FPS $\uparrow$ (f/s) & FLOPs $\downarrow$ (G)\\
        \midrule
        DCP~\cite{dcp} & TPAMI 2010 & 11.01 & 0.4165 & 13.28 & 0.4954 & 11.68 & 0.7090 & - & - \\ 
		
        DehazeNet~\cite{dehazenet} & TIP 2016 & 9.48 & 0.4383 & 16.62 & 0.5238 & 11.77 & 0.6217 & 1088.14 & 1.162 \\

        AOD-Net~\cite{aod} & ICCV 2017 & 13.14 & 0.4144 & 13.44 & 0.4136 & 12.33 & 0.6311 & 2564.10 & 0.230 \\	
        
        GCANet~\cite{gca} & WACV 2019 & 12.62 & 0.4208 & 17.49 & 0.5918 & 18.79 & 0.7729 & 270.64 & 36.82 \\

        \midrule

        GDNet~\cite{gdnet} & ICCV 2019 & 14.96 & 0.5326 & 13.80 & 0.5370 & 19.26 & 0.8046 & 100.96 & 42.98 \\	

        FFA-Net~\cite{ffa} & AAAI 2020 & 16.31 & 0.5362 & 18.60 & 0.6374 & 20.45 & 0.8043 & 17.89 & 575.6 \\
  
        MSBDN~\cite{msbdn} & CVPR 2020 & 15.13 & 0.5551 & 19.23 & 0.7056 & 20.11 & 0.8004 & 75.47 & 83.08 \\	
		
        AECR-Net~\cite{aecr} & CVPR 2021 & 15.80 & 0.4660 & 19.88 & \textbf{0.7173} & 20.68 & 0.8282 & 164.07 & 104.4 \\
  
        DeHamer~\cite{dehamer} & CVPR 2022 & 16.62 & 0.5602 & 20.66 & 0.6844 & 19.18 & 0.7939 & 2.34 & 59.62 \\
		
        SFNet~\cite{sfnet} & ICLR 2023 & 17.46 & 0.5780 & 16.90 & 0.7052 & 17.81 & 0.8291 & 12.80 & 125.4 \\	
  
        C2PNet~\cite{c2p} & CVPR 2023 & 16.88 & 0.5728 & - & - & 21.19 & 0.8334 & 0.30 & 461.2 \\
        
        \midrule
        
        FALCON & & \textbf{19.51} & \textbf{0.5860} & \textbf{20.84} & 0.6772 & \textbf{22.41} & \textbf{0.8357} & 182.90 & 57.27 \\
  
        \bottomrule
    \end{tabular}
    }
    \label{tab:PSNRSSIMcomparison}
\end{table}

\subsection{Dehazing Results}
\noindent\textbf{PSNR and SSIM.} \Cref{tab:PSNRSSIMcomparison} shows the dehazing quality of all the methods on all three datasets in PSNR and SSIM. We first observe that FALCON consistently surpasses other methods with the highest PSNR and SSIM across the datasets. (1) On Dense-Haze, FALCON shows the highest PSNR of \textbf{19.51 dB} and SSIM of \textbf{0.5860}. We further note that our PSNR shows a gain of +2.05 (+11.7\%) over the state-of-the-art method, which is significant considering the typical rate of improvement.  (2) On NH-Haze, FALCON also shows the highest PSNR of \textbf{20.84 dB}. (3) On NH-Haze2, FALCON achieves the best PSNR of \textbf{22.41 dB} and SSIM of \textbf{0.8357}.

\noindent\textbf{FPS and FLOPs.} In \Cref{tab:PSNRSSIMcomparison}, we show that FALCON achieves \textbf{182.90 FPS}, which means it can process over 180 images per second. This also equates 5 milliseconds inference time per image, which makes FALCON an extremely fast system, even applicable to standard 24 FPS videos in real-time. While there exist faster methods (i.e., rows 2 to 4 in \Cref{tab:PSNRSSIMcomparison}), their PSNR and SSIM significantly underperform compared to ours, making FALCON the fastest approach within the top 5 models in each dataset. We believe the \textit{best quality and speed} demonstrated by FALCON is a significant contribution towards standardizing practical image dehazing systems.

\begin{figure}[t]
	\centering
        \includegraphics[width=\linewidth]{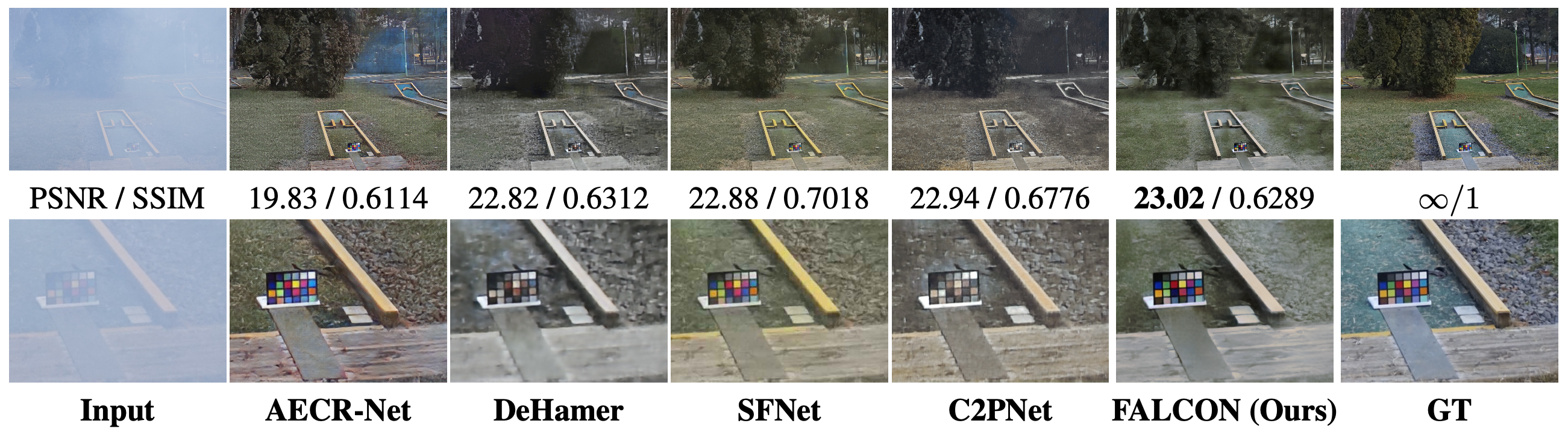}
	\caption{Comparative visualization of dehazing results on Dense-Haze. The top row displays the overall results, while the bottom row shows a magnified view.}
	\label{fig:Dense_Haze}
\end{figure}

\begin{figure}[t]
	\centering
        \includegraphics[width=\linewidth]{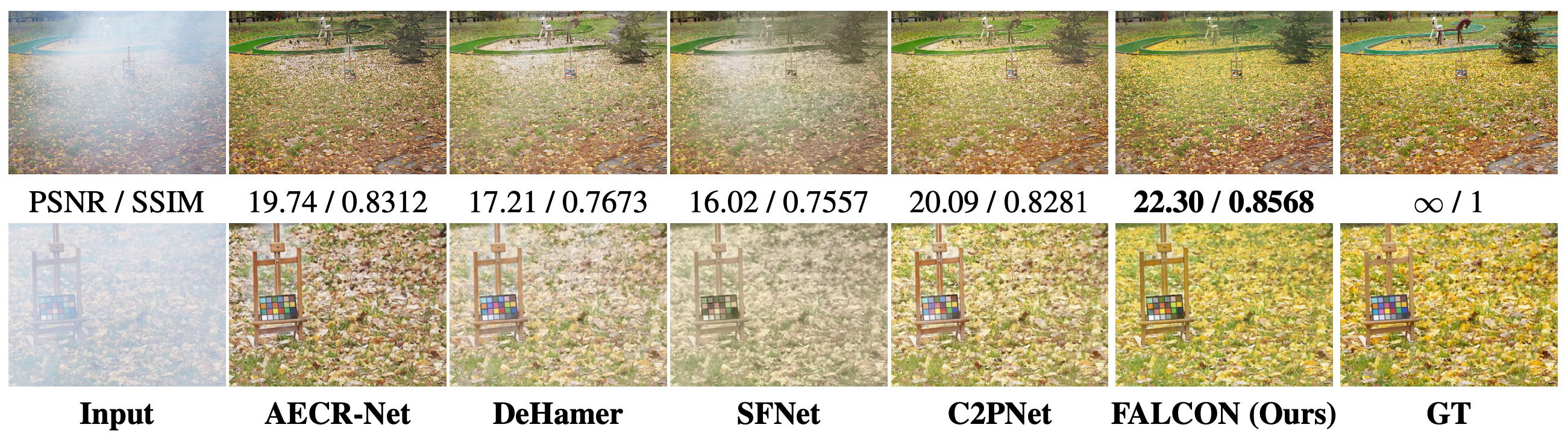}
	\caption{Comparative visualization of dehazing results using the NH-Haze2 dataset. The top row displays the overall results, while the bottom row shows a magnified view.}
	\label{fig:NH_Haze2}
\end{figure}

\noindent\textbf{Qualitative Results.} Based on the visual assessment shown in \cref{fig:Dense_Haze} and \cref{fig:NH_Haze2}, FALCON demonstrates a superior performance. On Dense-Haze, it eliminates haze while preserving the naturalness of the scene, sidestepping issues like color distortion and texture loss that plague other methods. Close-up views further underscore the details that FALCON retrieves, details that are often obscured in the outputs of other dehazing techniques. For NH-Haze2, FALCON reveals the underlying scene structures behind the dense haze, delivering results that are more true to the ground truth in both color and detail. In contrast, while AECR-Net~\cite{aecr} and SFNet~\cite{sfnet} may appear to produce visually pleasing results at first glance, a closer examination reveals significant color discrepancies with the ground truth. Certain areas exhibit completely unnatural color restoration, a flaw not present in the FALCON outputs, which maintains color fidelity throughout.

\subsection{Ablation Study}
\noindent\textbf{Quantitative Results.} In \Cref{tab:ablation}, we ablate FALCON to assess its performance under different combinations of FAL, CDM, and $\mathcal{L}_{map}$ on NH-Haze2. When none of them are included, FALCON is equivalent to U-Net. We first note that FAL alone brings a considerable improvement. Also, CDM significantly improves the performance. While $\mathcal{L}_{map}$ does provide some gain, $\mathcal{L}_{map}$, including all three components brings the best result.

\begin{figure}[t]
	\centering
        \includegraphics[width=\linewidth]{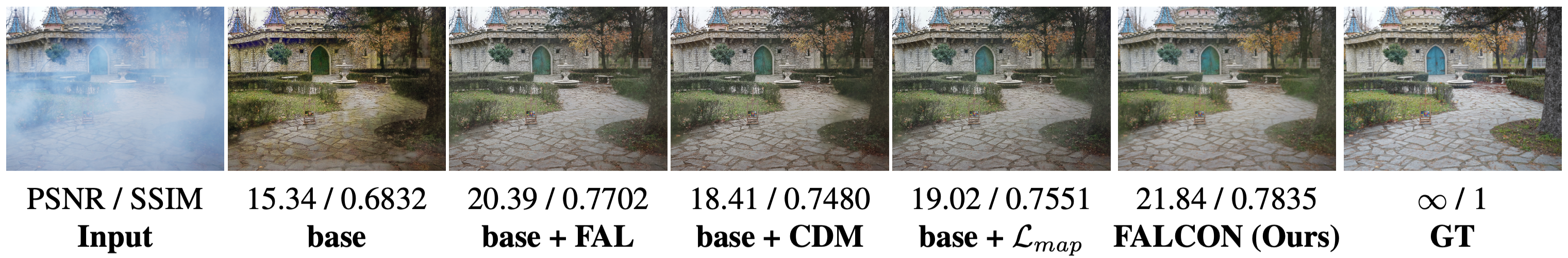}
	\caption{An ablation study showcasing the progressive enhancement of our method. From left to right: Input image, Base setting without our proposed enhancements, Base+FAL (Frequency Adjoint Link), Base+CDM (Continuous Density Mask), Base+Density Map Loss, our final method combining all enhancements, and Ground Truth (GT) image. Each column demonstrates the visual improvements achieved by incrementally integrating our proposed techniques.}
	\label{fig:ablation}
\end{figure}

\noindent\textbf{Qualitative Results.} \cref{fig:ablation} provides a visual comparison across the different setups shown in \Cref{tab:ablation}. The progressive improvement in image clarity and detail recovery is evident with each added component. Our full model exhibits substantial dehazing effectiveness, securing between detail preservation and haze removal. 

\begin{table}[t]
\caption{Ablation study on FALCON assessing the effects of including the Frequency Adjoint Link (FAL), Continuous Density Mask (CDM), and $\mathcal{L}_{map}$}
\centering
\begin{tabular}{c | c c c c c c c c}

\toprule
FAL & - & \checkmark & - & - & \checkmark & - & \checkmark & \checkmark \\
CDM & - & - & \checkmark & - & \checkmark & \checkmark & - & \checkmark \\
$\mathcal{L}_{map}$ & - & - & - & \checkmark & - & \checkmark & \checkmark & \checkmark \\
\midrule
PSNR $\uparrow$ & 17.76 & 18.93 & 19.72 & 17.81 & 20.21 & 19.93 & 20.82 & \textbf{22.41} \\
SSIM $\uparrow$ & 0.7644 & 0.7900 & 0.7975 & 0.7784 & 0.8012 & 0.8077 & 0.7963 & \textbf{0.8357} \\
\bottomrule

\end{tabular}
\label{tab:ablation}
\end{table}

\subsection{Computational Performance Analysis}
\noindent\textbf{Comparison Across Resolutions.} \Cref{tab:gpu5} shows the inference speed in FPS using various image sizes and GPUs. We observe that across varying GPUs, FALCON maintains over 100 FPS in almost all cases. Also, on GTX 1080 Ti, FALCON processes 256$\times$256 images at an impressive \textbf{249.42} frames per second, and even at a resolution of 2048$\times$2048, it sustains a rate of \textbf{46.51} frames per second, which is remarkable for such detailed imagery. Interestingly, using T4 which the Google Colab provides easily performs beyond 100 FPS.
The results indicate that FALCON achieves a frame rate well above the 30 frames per second threshold required for real-time processing, even on high-resolution images. 

\noindent\textbf{Comparison on GPUs.} Moreover, \Cref{tab:fpscompare} presents a comprehensive GPU performance comparison in FPS achieved by FALCON and other leading dehazing methods. The table illustrates FALCON's exceptional speed across various GPUs, including NVIDIA RTX 3090, GTX 1080 Ti, and  RTX A6000. While this table specifically addresses computational efficiency, it's important to refer to \Cref{tab:PSNRSSIMcomparison} for detailed dehazing quality metrics. Together, these results underscore FALCON's capability to deliver high-speed processing without compromising on the quality of dehazing, making it a highly practical choice for real-time image dehazing.
\begin{table}[t]
	\caption{Performance comparison of FPS (frames per second) across different image sizes and NVIDIA GPU models, demonstrating the scalability and efficiency of our method in various computational environments}
	\centering
	\begin{tabular}{c|c c c c}
		\toprule

            Image Size  & RTX A6000 & RTX 3090 & GTX 1080 Ti & T4 \\

            \midrule

            256$\times$256 & 147.48 & 182.90 & 249.42 & 156.80 \\
            512$\times$512 & 120.88 & 178.27 & 195.48 & 130.39 \\
            1024$\times$1024 & 115.93 & 161.15 & 124.22 & 121.68 \\
            2048$\times$2048 & 109.98 & 157.94 & 46.51 & 111.53 \\
  
		\bottomrule
	\end{tabular}
	\label{tab:gpu5}
\end{table} 

\begin{table}[t]
    \caption{GPU performance comparison in FPS across various methods. All FPS measurements were conducted based on images with a resolution of 1024$\times$1024, showing superior inference speed against other state-of-the-art methods across different NVIDIA GPUs. 
    }
    
    \centering 
    \begin{tabular}{c | c c c c}
    \toprule
     GPU      & DeHamer~\cite{dehamer} & SFNet~\cite{sfnet} & C2PNet~\cite{c2p} & FALCON  \\
     \midrule
    RTX A6000 & 38.96       &  2.45     & 1.14      & \textbf{115.93}        \\
    RTX3090      & 0.135       & 2.557     & 0.015      & \textbf{161.15}        \\
    GTX 1080 Ti  & 0.162       & 1.218     & 0.328      & \textbf{124.22}        \\

    \bottomrule
    \end{tabular}
    \label{tab:fpscompare}
\end{table}

\subsection{Versatility of Continuous Density Mask (CDM)}
\begin{figure}[t]
    \centering
    \begin{tabular}{c}
        \includegraphics[width=\linewidth]{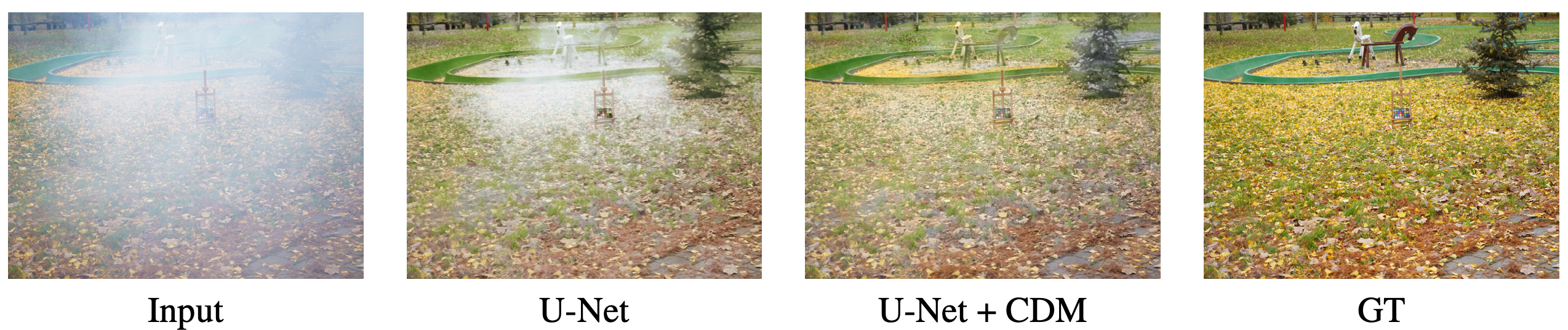} \\
        \\
        \includegraphics[width=\linewidth]{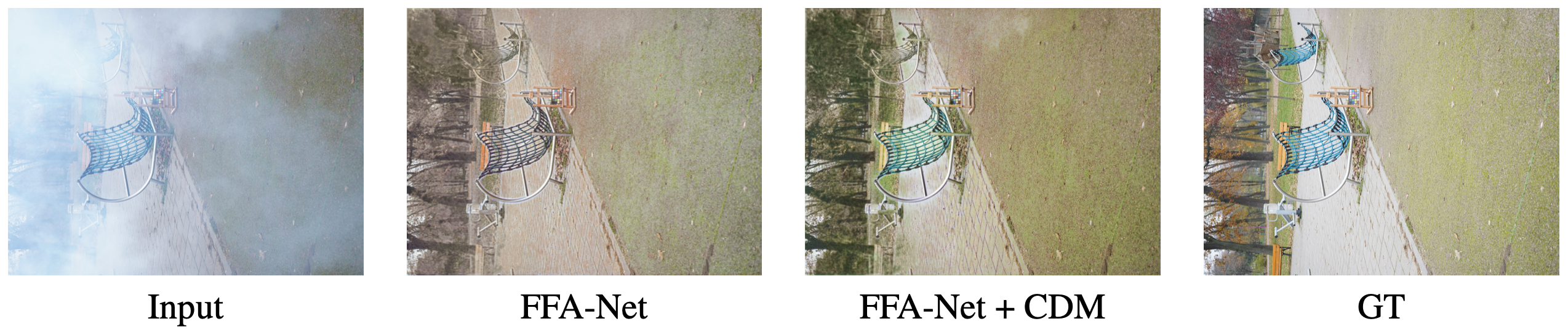} \\
    \end{tabular}
    \caption{Comparative visualization of dehazing results with and without CDM}
    \label{fig:cdm_total}
\end{figure}

\noindent To evaluate the effectiveness and versatility of the Continuous Density Mask (CDM), we designed experiments comparing the performance with and without the use of CDM across various networks. To accurately assess the effectiveness of CDM, we maintained identical settings for other hyperparameters, including learning rate and optimizer, and the experimental environment. To assess the versatility of CDM, we utilized widely used backbone networks in various fields, including U-Net~\cite{unet} and U-Net++~\cite{unet++}, as well as networks proposed specifically for the dehazing task, such as FFA-Net~\cite{ffa} and MSBDN~\cite{msbdn}. The dataset used for evaluation was NH-Haze2~\cite{nh-haze2}, with PSNR and SSIM as the metrics. During training, only the  Mean Squared Error (MSE) loss between images was used as the loss function. 
\Cref{tab:cdm_sup} presents the PSNR and SSIM results for each network, trained and evaluated with and without the application of CDM in the pre-network stage. It can be observed that the performance of all the various networks improved when CDM was applied before the network stage. \cref{fig:cdm_total} compares the dehazed images from each network with and without the application of CDM, confirming that the dehazed images with CDM applied demonstrate superior performance.
\begin{table}[t]
    \caption{Continuous Density Mask (CDM) on various networks}
    \centering
    \begin{tabular}{c | c c c c c c c c}
         \toprule
         Network & \multicolumn{2}{c}{U-Net~\cite{unet}} & \multicolumn{2}{c}{U-net++~\cite{unet++}} & \multicolumn{2}{c}{FFA-Net~\cite{ffa}} & \multicolumn{2}{c}{MSBDN~\cite{msbdn}} \\
         \cmidrule(lr){2-3} \cmidrule(lr){4-5} \cmidrule(lr){6-7} \cmidrule(lr){8-9}
         CDM & - & \checkmark & - & \checkmark & - & \checkmark & - & \checkmark \\
         \midrule
         PSNR $\uparrow$ & 18.01 & \textbf{18.68} & 16.94 & \textbf{16.97} & 18.06 & \textbf{18.96} & 18.21 & \textbf{19.05} \\
         SSIM $\uparrow$ & 0.7684 & \textbf{0.7935} & 0.6202 & \textbf{0.6209} & 0.7866 & \textbf{0.7987} & 0.7679 & \textbf{0.7792} \\
         \bottomrule
    \end{tabular}
    \label{tab:cdm_sup}
\end{table}

\section{Conclusion}
\label{section:conclusion}

In this study, we introduce the FALCON, a novel method for image dehazing that effectively balances high-quality dehazing with the demands of real-time applications. Our method, leveraging the innovative Frequency Adjoint Link, demonstrates a significant enhancement in handling high-resolution images through a wide receptive field, without the need for a complex network structure. The presentation of the Continuous Density Mask and Density Map Loss further elevates the network performance, enabling more precise and efficient dehazing by utilizing haze density information. While FALCON marks a substantial advancement in the field, it is noteworthy that the network is not the most lightweight in terms of parameters, suggesting potential areas for future optimization. Future work could focus on refining the network architecture to reduce its parameter count while maintaining, or even enhancing its current performance levels. Overall, FALCON represents a promising step forward in real-time image dehazing, offering a blend of efficiency and effectiveness that is well-suited for critical applications in autonomous driving, surveillance, and national defense technologies.


%
%
\bibliographystyle{splncs04}
\bibliography{main}

\newpage

\alphaSection{Dataset Details}
\label{sec:datasets}
\begin{table}[h] 
    \caption{Details of datasets used in the image dehazing}
    \centering
    \scalebox{0.72}{
    \begin{tabular}{c | c c c | c c}

        \toprule

        Haze generation & \multicolumn{3}{c | }{\textbf{REAL-WORLD}} & \multicolumn{2}{c}{\textbf{SYNTHETIC}} \\
        Dataset & \textbf{Dense-Haze} & \textbf{NH-Haze} & \textbf{NH-Haze2} & \textbf{SOTS-indoor} & \textbf{SOTS-outdoor} \\
        Scene & {Out/indoor} & {Outdoor} & {Outdoor} & {Indoor} & {Outdoor} \\
        Haze type & Homogeneous & Non-homogeneous & Non-homogeneous & Homogeneous & Homogeneous \\
        Haze density & \textcolor{red}{Dense} & \textcolor{orange}{Moderate} & \textcolor{orange}{Moderate} & \textcolor{cyan}{Low} & \textcolor{cyan}{Low} \\

        \midrule

        DW-GAN~\cite{dwgan} (CVPRW'21)     & \checkmark & \checkmark & \checkmark & \checkmark & - \\
        AECR-Net~\cite{aecr} (CVPR'21)   & \checkmark & \checkmark & -          & \checkmark & - \\
        DeHamer~\cite{dehamer} (CVPR'22)         & \checkmark & \checkmark & -          & \checkmark & \checkmark \\
        PMDNet~\cite{pmdnet} (ECCV'22)      & \checkmark & \checkmark & -           & \checkmark & \checkmark \\
        FSDGN~\cite{fsdgn} (ECCV'22)  & \checkmark & \checkmark & - & \checkmark & - \\
        SFNet~\cite{sfnet} (ICLR'23)      & \checkmark & -          & -          & \checkmark & \checkmark \\
        C2PNet~\cite{c2p} (CVPR'23)     & \checkmark & -          & \checkmark & \checkmark & \checkmark \\
        Fourmer~\cite{fourmer} (ICML'23)    & \checkmark & \checkmark & -          & \checkmark & -         \\
        
        \midrule

        \textbf{FALCON} & \checkmark & \checkmark & \checkmark & \checkmark & -           \\
        
        \bottomrule
    
    \end{tabular}
    }
    \label{tab:datasets}
\end{table}
\noindent we have utilized the best quality real-world datasets currently available in the dehazing research field. As shown in \Cref{tab:datasets}, we show the nature of dehazing datasets used by recent papers published at top venues.

\alphaSection{Detailed Information about the Fast Fourier Convolution Block (FFCB)}

\begin{figure}[t]
        \centering
	\includegraphics[width=.3\linewidth]{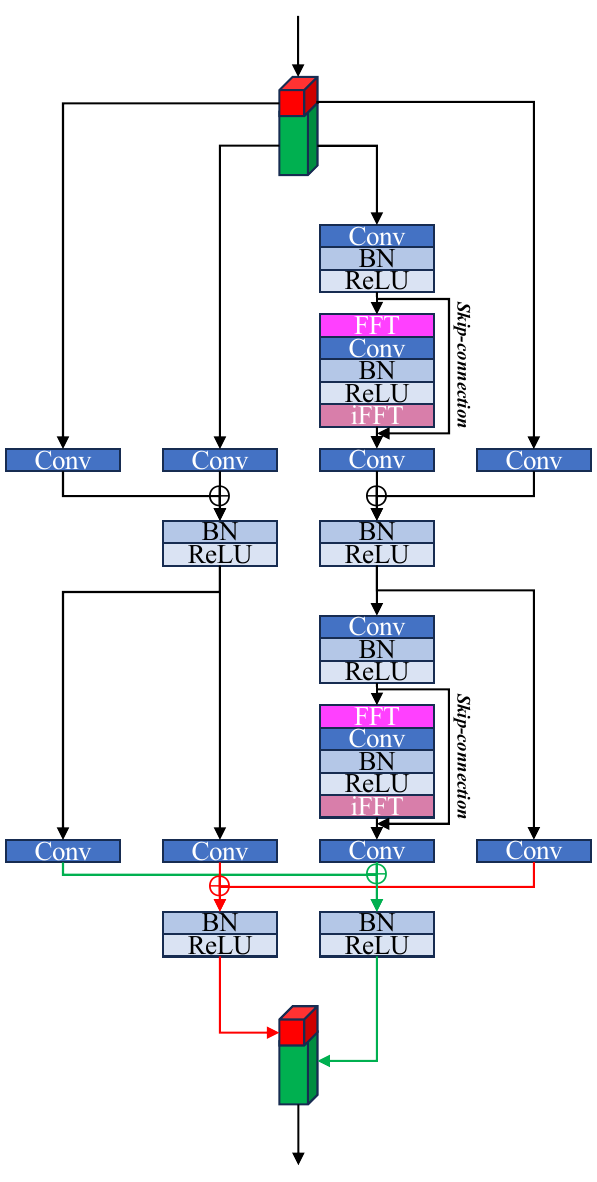}
	\captionsetup{justification=justified}
	\caption{Detailed Architecture of the Fast Fourier Convolution Block. The block diagram shows the detailed layer-wise composition, emphasizing the integration of Fourier transformations within the convolutional process. Each layer is annotated with its specific function, providing a comprehensive view of the block's operations. The symbol $\bigoplus$ denotes element-wise addition in this context. This design facilitates efficient dehazing by leveraging the frequency domain for enhanced receptive fields.}
	\label{fig:FFCBdetail}
\end{figure}

The main component of FALCON's network, the Frequency Adjoint Link (FAL), comprises elements such as Fast Fourier Convolution Block (FFCB) and convolution block. Among these, the FFCB effectively handles both local and global features of an image through various receptive fields. The feature \(\textbf{F}\) entering the FFCB first undergoes channel-wise partitioning. After extensive experimentation to find the optimal partition ratio, FALCON uses a 1:3 ratio for channel-wise partitioning. The two partitioned features are denoted as \(\textbf{F}^l\) and \(\textbf{F}^g\). \(\textbf{F}^l\) generates two features \(\textbf{F}^l_{s1}\) and \(\textbf{F}^l_{s2}\) through two convolutions in the spatial domain, while \(\textbf{F}^g\) produces the feature \(\textbf{F}^g_{f}\) through convolution in the frequency domain using Fast Fourier Transform (FFT) and the feature \(\textbf{F}^g_{s}\) through convolution in the spatial domain. The features \(\textbf{F}^l_{s1}\) and \(\textbf{F}^g_{s}\) are combined through element-wise addition, followed by batch normalization and ReLU to form the feature \(\textbf{F}^l_{mix1}\). Similarly, \(\textbf{F}^l_{s2}\) and \(\textbf{F}^g_{f}\) are combined through element-wise addition and then passed through batch normalization and ReLU in sequence to form the feature \(\textbf{F}^g_{mix1}\). Let's denote the network structures that derive \(\textbf{F}^l_{mix1}\) and \(\textbf{F}^g_{mix1}\) from \(\textbf{F}^l\) and \(\textbf{F}^g\) as \(local(\cdot)\) and \(global(\cdot)\), respectively. The operations up to this point can be expressed as follows:

\begin{equation} 
\textbf{F}^l_{mix1} = local(\textbf{F}^l, \textbf{F}^g)
\end{equation}
\begin{equation}
\textbf{F}^g_{mix1} = global(\textbf{F}^l, \textbf{F}^g).
\end{equation}

The next step involves using \(local(\cdot)\) and \(global(\cdot)\) to compute the subsequent features from \(\textbf{F}^l_{mix1}\) and \(\textbf{F}^g_{mix1}\). This process can be expressed as

\begin{equation}
\textbf{F}^l_{mix2} = local(\textbf{F}^l_{mix1}, \textbf{F}^g_{mix1})
\end{equation}
\begin{equation}
\textbf{F}^g_{mix2} = global(\textbf{F}^l_{mix1}, \textbf{F}^g_{mix1}).
\end{equation}

The two resulting features, \(\textbf{F}^l_{mix2}\) and \(\textbf{F}^g_{mix2}\), are then combined through channel-wise concatenation to form the final output feature of the FFCB. The detailed structure of the FFCB can be seen in \cref{fig:FFCBdetail}.

\alphaSection{Results on Synthetic Datasets}

\begin{table}[h]
    \caption{Quantitative comparisons against other methods on synthetic datasets. FPS and FLOPs are based on the inference using an RTX 3090 GPU with images of 256x256 resolution.}
\def\arraystretch{1.1}%
    \centering
    \small
    \begin{tabular}{c c c c c c}
        \toprule
        \multirow{2}{*}{\centering Method} & \multirow{2}{*}{\centering Venue} & \multicolumn{2}{c}{SOTS} & \multicolumn{2}{c}{Computational Efficiency} \\
        
        \cmidrule(lr){3-4} \cmidrule(lr){5-6} 
        
        & & PSNR $\uparrow$ & SSIM $\uparrow$ & FPS $\uparrow$ (f/s) & FLOPs $\downarrow$ (G)\\
        \midrule\midrule
        DCP~\cite{dcp} & TPAMI 2010 & 16.62 & 0.8179 & - & - \\ 
		
        DehazeNet~\cite{dehazenet} & TIP 2016 & 19.82 & 0.821 & 1088.14 & 1.162 \\

        AOD-Net~\cite{aod} & ICCV 2017 & 20.51 & 0.816 & 2564.10 & 0.230 \\	
        
        GCANet~\cite{gca} & WACV 2019 & 30.06 & 0.9596 & 270.64 & 36.82 \\

        \midrule

        GDNet~\cite{gdnet} & ICCV 2019 & 32.16 & 0.9836 & 100.96 & 42.98 \\	

        FFA-Net~\cite{ffa} & AAAI 2020 & 36.39 & 0.9886 & 17.89 & 575.6 \\
  
        MSBDN~\cite{msbdn} & CVPR 2020 & 33.67 & 0.985 & 75.47 & 83.08 \\	
		
        AECR-Net~\cite{aecr} & CVPR 2021 & 37.17 & 0.9901 & 164.07 & 104.4 \\
  
        DeHamer~\cite{dehamer} & CVPR 2022 & 36.63 & 0.9881 & 2.34 & 59.62 \\
		
        SFNet~\cite{sfnet} & ICLR 2023 & 41.24 & 0.996 & 12.80 & 125.4 \\	
  
        C2PNet~\cite{c2p} & CVPR 2023 & 42.56 & 0.9954 & 0.30 & 461.2 \\
        
        \midrule
        
        \textbf{FALCON} & & 37.01 & 0.9762 & 182.90 & 57.27 \\
  
        \bottomrule
    \end{tabular}
    \label{tab:PSNRSSIMcomparison2}
\end{table}

\noindent Just as we measure PSNR and SSIM for real-world datasets, we also evaluate the performance on synthetic dataset. We use the RESIDE~\cite{reside} dataset to assess the performance of FALCON on synthetic data. Following the settings used in previous studies~\cite{gdnet, ffa, aecr}, we select ITS as the training dataset and SOTS-indoor for testing dataset. ITS and SOTS-indoor consist of 13,990 and 500 images, respectively. All other experimental settings were identical to those used for the real-world datasets. As can be seen in the results of \Cref{tab:PSNRSSIMcomparison2}, FALCON demonstrated competitive performance with a PSNR of 37.01dB and an SSIM of 0.9762, not falling behind other methods. 
\alphaSection{Additional Visual Results}

\begin{figure}
    \centering
    \includegraphics[width=\linewidth]{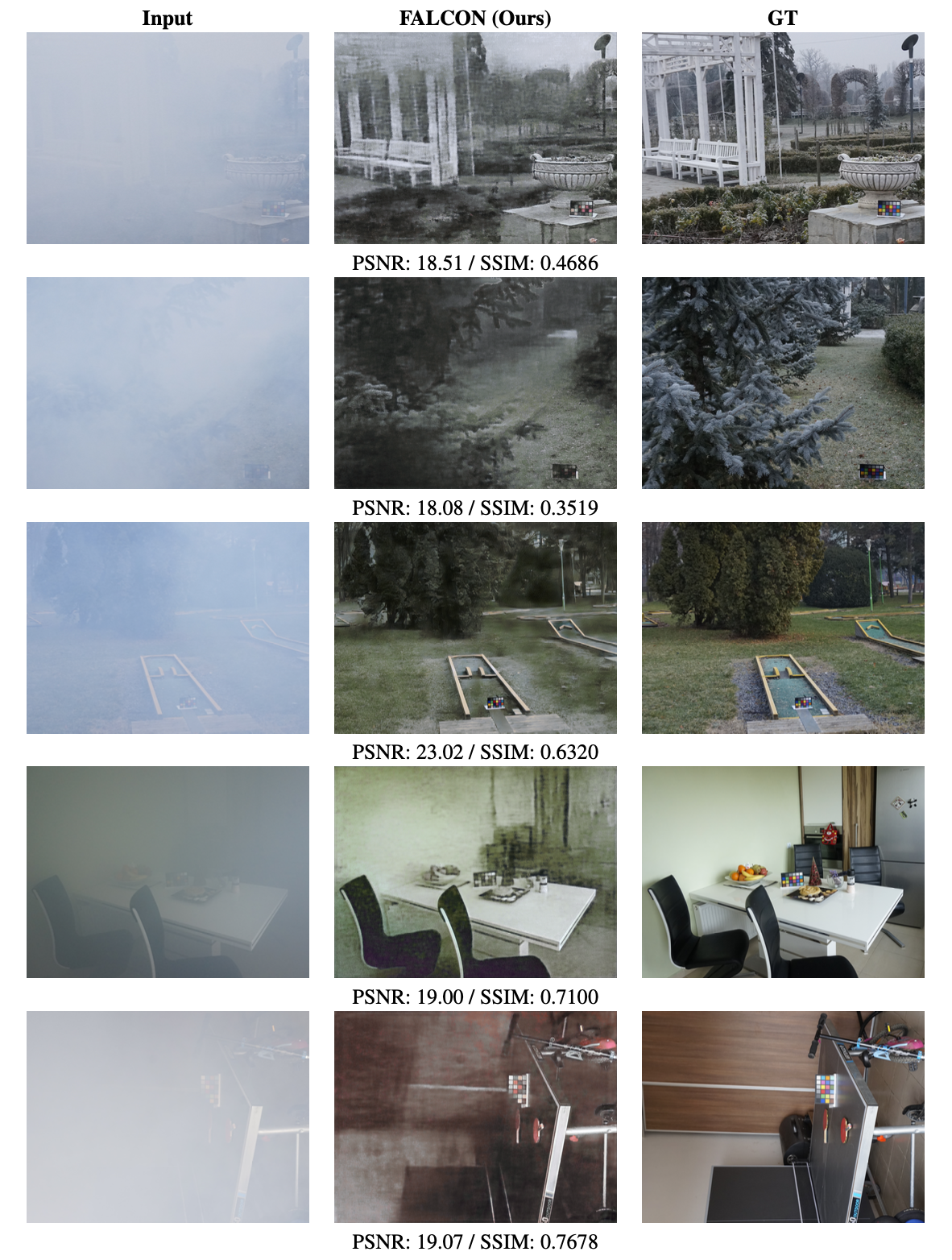}
    \caption{Dehazing results of FALCON on Dense-Haze.}
    \label{fig:alldense}
\end{figure}

\begin{figure}
    \centering
    \includegraphics[width=\linewidth]{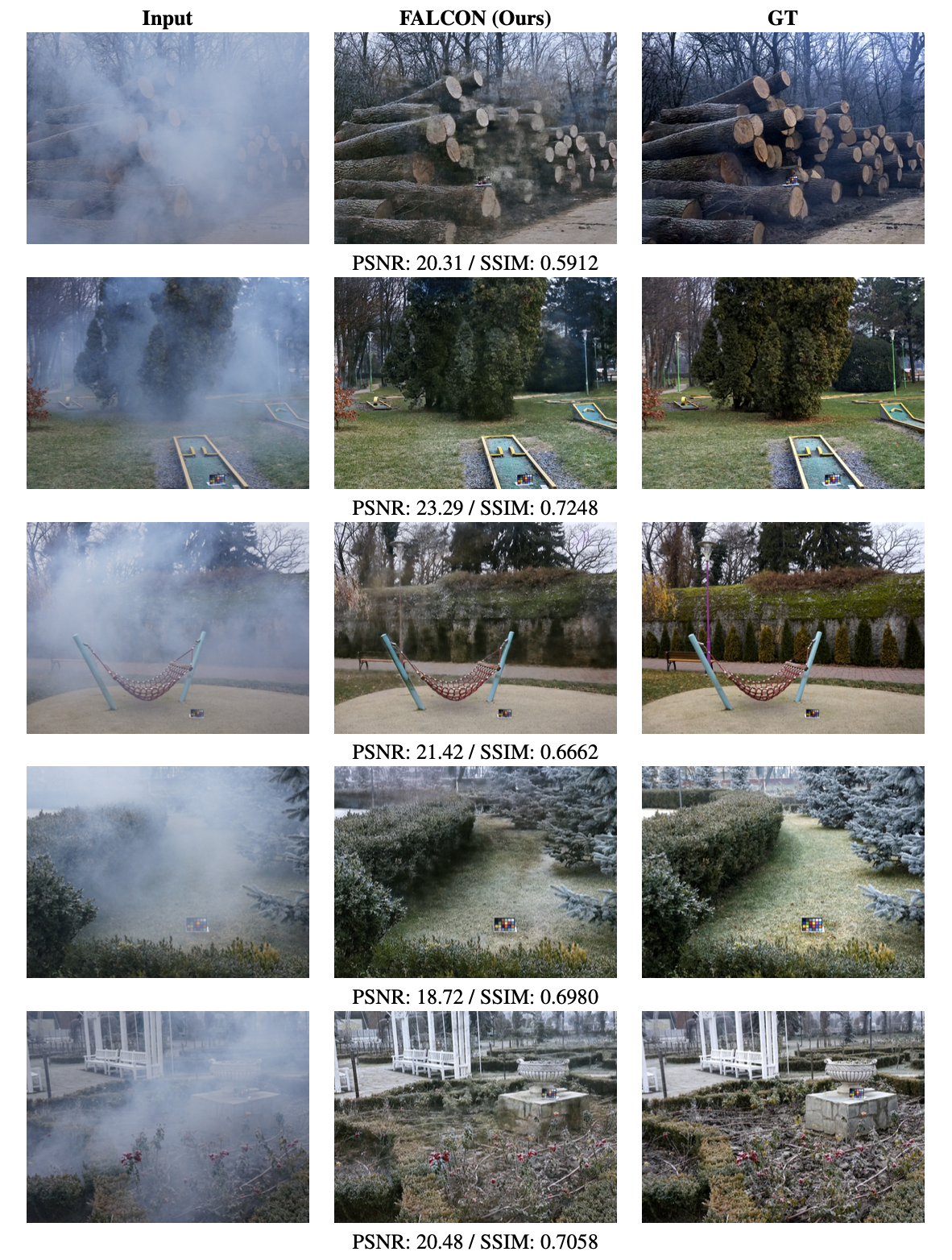}
    \caption{Dehazing results of FALCON on NH-Haze.}
    \label{fig:allnh1}
\end{figure}

\begin{figure}
    \centering
    \includegraphics[width=\linewidth]{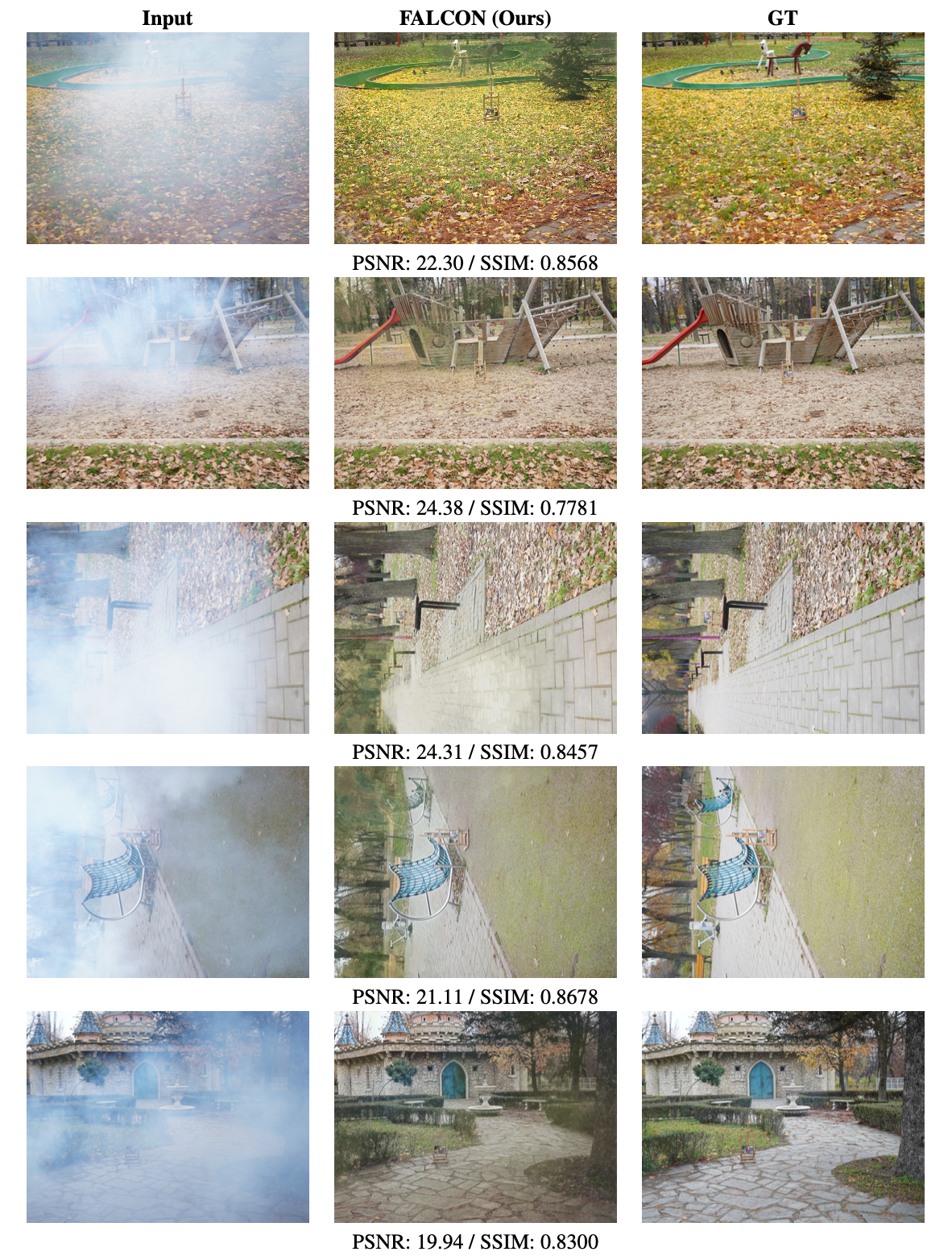}
    \caption{Dehazing results of FALCON on NH-Haze2.}
    \label{fig:allnh2}
\end{figure}

To accurately and transparently showcase the dehazing performance of FALCON, we present the results for test dataset images of every real-world dataset we experimented with. \cref{fig:alldense,fig:allnh1,fig:allnh2} display the outcomes of FALCON for images in the test datasets of Dense-Haze~\cite{dense-haze}, NH-Haze~\cite{nh-haze}, and NH-Haze2~\cite{nh-haze2}, respectively.

\alphaSection{Discussion}

\subsection{Limitation and Future Works}
Designed for real-time dehazing, FALCON demonstrates a commendable balance between rapid inference speed and high-quality dehazing. To achieve swift inference, the network architecture is kept as simple as possible, resulting in remarkably low FLOPs. Despite this, due to its U-net~\cite{unet} based structure, FALCON possesses a relatively high number of parameters. The parameter count of the FALCON network implies a substantial memory requirement. This aspect suggests that for successful implementation of real-time dehazing in edge devices, strategies to reduce the number of parameters should be considered.

Moreover, while FALCON is designed to be optimized for real-world dehazing scenarios, it is observed that its performance on synthetic datasets is not at the forefront. This indicates a potential area for improvement. Enhancing FALCON's adaptability to various datasets, especially synthetic ones, could further solidify its applicability across a broader range of dehazing tasks, ensuring robust performance irrespective of the dataset's origin.

\subsection{Broader Impact}
FALCON is a remarkable dehazing technique, demonstrating its potential for significant application in fields where dehazing is crucial. Despite having room for further development, its current capabilities make it sufficiently robust for use in real-world scenarios where dehazing is essential. The field of dehazing research is driven by the need for rapid dehazing techniques in applications such as autonomous driving, CCTV, and national defense technology. These applications require the ability to continuously capture hazy environments and produce clear images promptly for subsequent high-level computer vision tasks or other necessary operations. Thus, the ultimate goal of dehazing methods extends beyond merely improving the quality of dehazing; they must also be efficient and fast for real-time applications.

Given its exceptional dehazing quality and remarkably fast inference speed on real-world datasets, FALCON emerges as a highly suitable solution in the realm of dehazing technology. FALCON's capabilities extend beyond technical excellence, offering broader societal impacts and positive contributions to various sectors. Its proficiency in delivering high-quality dehazing and rapid inference speed on real-world datasets positions it as a pivotal tool in enhancing the effectiveness of critical applications such as autonomous driving, CCTV surveillance, and national defense. By ensuring clearer visual information in challenging atmospheric conditions, FALCON can significantly improve safety and reliability in these areas. Its application in autonomous driving, for instance, could lead to safer navigation and reduced accidents in poor visibility conditions. In surveillance and defense, clearer images can enhance monitoring accuracy and situational awareness. Thus, FALCON not only represents a technological advancement in the field of dehazing but also holds the potential to bring about substantial societal benefits by improving safety, security, and operational efficiency in various real-world scenarios.

\end{document}